\documentclass[letterpaper, 10 pt, conference]{ieeeconf}  % Comment this line out if you need a4paper

\IEEEoverridecommandlockouts                              % This command is only needed if 
\usepackage{amsmath,amssymb,amsfonts}
\usepackage{graphicx} % for pdf, bitmapped graphics files
\usepackage[dvipsnames]{xcolor}
\usepackage{algorithm}
\usepackage{booktabs} 
\usepackage{algpseudocode}
\usepackage{multirow}
\usepackage{svg}
\usepackage{xspace}
\usepackage{xcolor}
\usepackage{colortbl}
\usepackage{tikz}
\usepackage{pifont}
\usepackage{gensymb}

\colorlet{tabfirst}{Green!35}
\definecolor{tabthird}{rgb}{1, 0.85, 0.7}
\definecolor{tabsecond}{rgb}{1, 0.96, 0.7}

\newcommand{\ours}{\text{COLMAR}\xspace}

\makeatletter
\let\NAT@parse\undefined
\makeatother
\usepackage{hyperref}
\hypersetup{
    colorlinks=true,
    linkcolor=blue,
    filecolor=magenta,      
    urlcolor=cyan,
    citecolor=red,
    pdftitle={COLMAR},
    pdfpagemode=FullScreen,
}

\title{\LARGE \bf
\ours: Cooperative View Policy Learning for Multi-Agent Active 3D Reconstruction
}

\author{Phu Pham$^{1}$, Damon Conover$^{2}$,  Aniket Bera$^{1}$\\
$^1$Department of Computer Science, Purdue University  $^2$DEVCOM Army Research Laboratory\\
\texttt{\{phupham, aniketbera\}@purdue.edu, damon.m.conover.civ@army.mil}
}

\begin{document}

\maketitle
\thispagestyle{empty}
\pagestyle{empty}

%%%%%%%%%%%%%%%%%%%%%%%%%%%%%%%%%%%%%%%%%%%%%%%%%%%%%%%%%%%%%%%%%%%%%%%%%%%%%%%%
\begin{abstract}
Active 3D reconstruction requires selecting informative viewpoints under limited sensing budgets. In multi-agent settings, coordination inefficiencies such as redundant observations and spatial clustering can significantly reduce reconstruction quality. We present COLMAR, a cooperative view policy learning framework for multi-agent active 3D reconstruction. COLMAR formulates viewpoint allocation as a shared policy optimization over map-centric observations and introduces a reconstruction-aware objective that promotes overlap-aware coverage, team-level discovery, and collision-safe exploration. Dense feedback derived from incremental reconstruction updates aligns exploration behavior with downstream geometric quality. The policy is trained using parameter-sharing Proximal Policy Optimization (PPO) with independent per-agent action selection at deployment, conditioned on a fused team map and without inter-agent message passing for decision making. Selected viewpoints are then reconstructed with 3D Gaussian Splatting (3DGS) for high-fidelity photometric evaluation. Experiments on GLEAM and Replica demonstrate consistent improvements over heuristic and non-cooperative baselines, achieving up to 54\% higher reconstruction accuracy and 49\% greater coverage under matched sensing budgets.
\end{abstract}

%%%%%%%%%%%%%%%%%%%%%%%%%%%%%%%%%%%%%%%%%%%%%%%%%%%%%%%%%%%%%%%%%%%%%%%%%%%%%%%%
\section{INTRODUCTION}
\label{sec:intro}
Autonomous 3D reconstruction has become a core capability for inspection, digital twinning, search-and-rescue, and embodied intelligence. The active variant of this problem requires a robot to decide where to observe next so that a fixed sensing budget yields maximal reconstruction quality. Classical solutions are typically built on frontier/utility heuristics \cite{yamauchi1997frontier} and next-best-view (NBV) criteria \cite{bircher2016receding}. While these methods are practical and robust, they are often brittle under complex geometry, partial observability, and nontrivial motion constraints, where handcrafted scoring terms can produce myopic or redundant trajectories.

Recent learning-based active reconstruction methods have shown that policy learning can outperform fixed NBV heuristics in difficult scenes. For example, GenNBV \cite{chen2024gennbv} learns a generalizable reinforcement learning (RL) policy for active reconstruction in free-space actions. GLEAM \cite{chen2025gleam} further improves generalization for active mapping in complex 3D indoor scenes. In parallel, reconstruction backends have advanced rapidly, from neural implicit simultaneous localization and mapping (SLAM) \cite{zhu2022niceslam, nerfslam} to efficient Gaussian representations \cite{kerbl2023gaussiansplatting}. Recent multi-agent SLAM systems such as MNE-SLAM \cite{mneslam2025, magicslam2025, grandslam2025} further highlight this trend. Active reconstruction systems such as ActiveGAMER \cite{chen2025activegamer} also strengthen the case for tighter integration between mapping state and cooperative view planning.

Despite this progress, two gaps remain for multi-agent active 3D reconstruction. First, many approaches still optimize per-agent exploration with limited coordination signals, leading to duplicated observations and inefficient team-level coverage \cite{golodetz2018collaborative,xu2024macego3d}. Second, policy learning is often supervised only by final or coarse rewards, whereas rich reconstruction diagnostics, such as local coverage growth, under-explored-region cues, and motion validity, are not fully leveraged during training \cite{chen2024gennbv,chen2025activegamer}. As a result, coordination quality and policy stability can degrade as scene complexity and team size increase.

\begin{figure}[t]
    \centering
    \includegraphics[width=\linewidth]{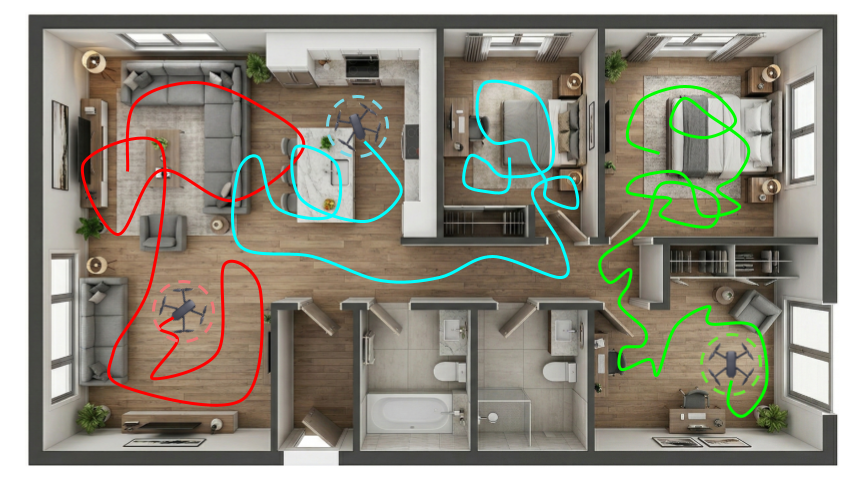}
    \caption{Illustration of \ours for cooperative multi-agent active 3D reconstruction. Agents leverage a shared map context to coordinate view selection, reducing redundancy and improving reconstruction quality under fixed sensing budgets.}
    \label{fig:teaser}
\end{figure}

To address these issues, we present \ours, a cooperative view policy learning framework for multi-agent active 3D reconstruction. \ours trains a shared map-aware policy with reconstruction-driven rewards that explicitly encourages informative, complementary, and safe view selection. The policy is trained with truncated signed distance function (TSDF) mapping for occupancy, frontier, and reward signals. 3D Gaussian Splatting (3DGS) is used only afterward for photorealistic reconstruction and rendering-based evaluation, and does not affect the learned view policy. As illustrated in Fig. \ref{fig:teaser}, \ours enables agents to leverage a shared map context to coordinate view selection, reducing overlap while improving reconstruction quality.

The main contributions are:
\begin{itemize}
    \item We formulate multi-agent active 3D reconstruction as cooperative policy learning over map-centric observations, enabling agents to coordinate viewpoint allocation through a shared policy.
    \item We introduce a reconstruction-aware objective that combines policy optimization with reward components tied to local coverage gain, under-explored-region discovery, and collision-aware behavior.
    \item We provide a scalable training and evaluation pipeline and show that \ours improves reconstruction quality and exploration efficiency against strong heuristic and learning-based baselines under matched sensing budgets.
\end{itemize}

\begin{figure*}
    \centering
    \includegraphics[width=\linewidth]{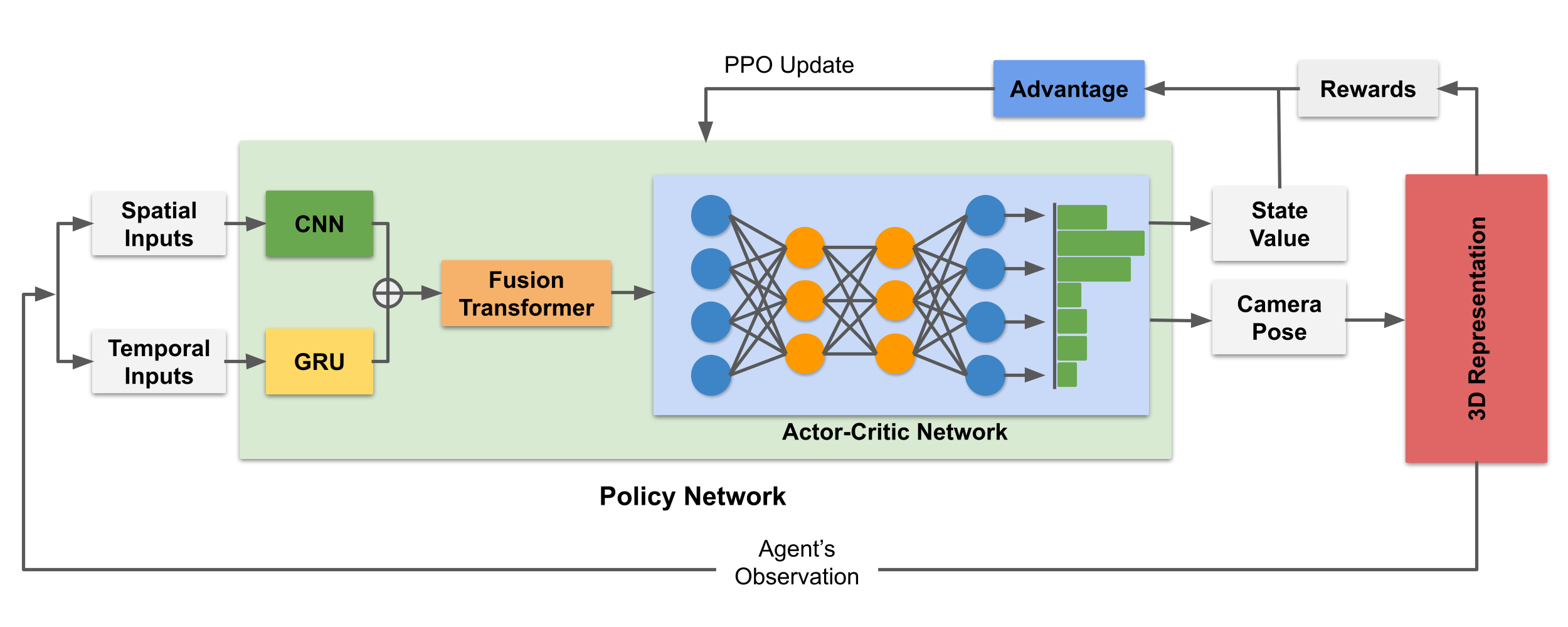}
    \caption{Overview of COLMAR. During training, a shared TSDF provides map-centric observations and reconstruction-aware rewards for PPO. During deployment, the learned policy selects views independently per agent. Selected trajectories are reconstructed with 3DGS for photorealistic evaluation. 3DGS is not used to train or update the policy.}
    \label{fig:overview}
\end{figure*}

\section{RELATED WORK}
\label{sec:related}
\subsection{Active View Planning and Next-Best-View}
Classical exploration and NBV methods are rooted in frontier expansion and utility-based scoring, where candidate views are selected to maximize expected map gain while satisfying geometric constraints \cite{yamauchi1997frontier,bircher2016receding}. Recent robotics work has improved online NBV planning for practical platforms, including mobile manipulators and aerial systems \cite{naazare2022online,batinovic2022shadowcasting}. These methods are effective and interpretable, but their hand-crafted criteria and planner-specific assumptions can limit adaptability across scene types, sensor regimes, and action spaces.

\subsection{Multi-Agent Exploration and Coordination}
Multi-robot exploration research has long emphasized task allocation, redundancy reduction, and communication-aware coordination \cite{burgard2005coordinated, cramp, patel2023dream}. In practice, however, many systems still rely on decoupled per-agent utility assignment, which can underutilize shared map context when the team size grows or visibility is highly nonuniform. Collaborative dense mapping systems with online inter-agent pose optimization \cite{golodetz2018collaborative} showed that consistent global fusion across agents is feasible at building scale. More recent cooperative systems such as MAC-Ego3D \cite{xu2024macego3d} and CORE \cite{wang2023core} demonstrate stronger team coordination through map sharing and coordination-aware planning. These results highlight both the opportunity and the challenge of learning coordinated view allocation policies that improve exploration efficiency and reconstruction fidelity.

\subsection{Learning-Based Active Reconstruction and Mapping}
Learning-based active exploration has shown that policy optimization can outperform static heuristics in complex environments. Active Neural SLAM \cite{chaplot2020activeneuralslam} demonstrated this direction in map-based embodied exploration. For active mapping and reconstruction, GenNBV \cite{chen2024gennbv} introduced a generalizable RL policy with freer action parameterization, while GLEAM \cite{chen2025gleam} emphasized broad cross-scene generalization in complex indoor layouts. In parallel, reconstruction backbones have advanced from neural implicit mapping (e.g., NICE-SLAM \cite{zhu2022niceslam}) to efficient 3D Gaussian representations \cite{kerbl2023gaussiansplatting, monogs, pham2024flashslam}. Multi-agent systems including MNE-SLAM \cite{mneslam2025}, MAGiC-SLAM \cite{magicslam2025}, and GRAND-SLAM \cite{grandslam2025} show strong progress in consistency and rendering quality. Still, these advances do not fully resolve coordinated multi-agent view allocation under reconstruction-driven objectives.

\subsection{Training Paradigms for Multi-Agent Policy Learning}
Cooperative multi-agent RL is commonly instantiated with centralized training and decentralized execution (CTDE) actor-critic methods (e.g., MADDPG \cite{lowe2017maddpg}, MAPPO \cite{lan2025mappo, yu2022mappo}), value-decomposition methods (e.g., QMIX \cite{rashid2018qmix}), or decentralized parameter-sharing policies. In parallel, large-scale distributed policy optimization has been shown to improve sample throughput and stability via actor-learner decoupling and synchronized updates \cite{espeholt2018impala}. CTDE can improve credit assignment, but it usually requires additional centralized inputs and training machinery. Value-decomposition methods are powerful, but they are often tied to factored value assumptions and discrete coordination structure. We instead adopt cooperative parameter-sharing PPO with decentralized execution and distributed optimization, which matches our setting where agents share motion and observation structure and must scale across many parallel reconstruction episodes.

Unlike prior approaches that either optimize heuristic coordination or treat mapping feedback as a weak training signal, \ours is designed around a reconstruction-native learning objective: map-centric policy inputs and explicit reconstruction-aware rewards aligned with coordination-critical behaviors (local coverage progress, under-explored-region preference, and collision-aware motion). Our main contribution is the specific reconstruction-aligned cooperative formulation. It combines overlap-aware unique coverage, TSDF-derived view-space gain, and a practical training/inference split (geometry-focused RL training, photorealistic 3DGS inference).

\section{METHOD}
\label{sec:method}
\subsection{Problem Formulation}
\label{subsec:problem}
\paragraph{Environment and Agents}
We consider active reconstruction in a 3D indoor scene $s \in \mathcal{S}$ with a team of $N$ agents, $\mathcal{A}=\{1,\dots,N\}$. At time step $t$, each agent $i$ receives a local observation $o_t^i$ and executes an action $a_t^i$. The action is selected from a discrete motion primitive set with translation along $x$, $y$, and $z$, and rotation in yaw and pitch. Roll is disabled. The joint action is $\mathbf{a}_t=(a_t^1,\dots,a_t^N)$. Episodes terminate at horizon $T$ or when an environment-specific stopping condition is met.

\paragraph{State and Mapping Context}
Each observation is map-centric and contains both perception and mapping context. In our implementation, the policy input for agent $i$ includes current pose, rendered depth, ego-centric occupancy map, and recent pose history. A mapping backend maintains a shared scene representation fused from all agents' measurements. During training, this representation is a shared TSDF volume held by the mapping backend. It supplies occupancy and frontier features in each agent's observation and provides the reconstruction-aware reward for PPO. During deployment, policy parameters are frozen. Agents continue to observe map-centric occupancy and frontier context derived from the fused team map and select actions independently. Collected viewpoints are then used for final 3DGS reconstruction and photometric evaluation. Decentralization therefore refers to decision making, not the absence of map fusion. Coordination arises from shared map evolution and shared policy parameters.

\subsection{Architecture Overview}

Figure~\ref{fig:overview} summarizes the overall architecture. During training, multiple agents collect map-centric observations from a shared TSDF maintained by the mapping backend, a shared policy predicts cooperative view actions, and TSDF updates yield reconstruction-aware rewards for PPO. During deployment, the frozen policy performs independent per-agent view planning from the same map-centric observation interface, and the resulting trajectories are used for final 3DGS reconstruction and rendering-based evaluation.

\subsection{Policy Input and Network Architecture}
We structure the policy input into \emph{spatial} and \emph{temporal} modalities. For agent \(i\), the spatial input contains rendered depth, ego-centric occupancy map, and a global map that summarizes visited/free/occupied/frontier information. The temporal input contains the current pose descriptor and a short history of recent poses.

The network uses two encoding branches. The spatial tensors are channel-wise stacked and encoded by a convolutional neural network (CNN) into a spatial latent \(h_{\text{sp}}^i\). The pose history is encoded by a gated recurrent unit (GRU) into a trajectory latent, which is concatenated with the current pose and mapped by a multi-layer perceptron (MLP) to a vector latent \(h_{\text{vec}}^i\). This yields

\begin{equation}
    h_{\text{sp}}^i = \mathrm{CNN}(x_{\text{sp}}^i), \qquad
    h_{\text{vec}}^i = \mathrm{GRU}(x_{\text{temp}}^i)
\end{equation}

We then fuse both latents with a lightweight token-fusion module. In the default setting, a transformer encoder performs self-attention over modality tokens, and then projects to a shared representation \(h^i\). This representation is consumed by actor-critic heads:
\begin{equation}
\mathbf{z}_t^i = A(h^i), \qquad V_t^i = C(h^i),
\end{equation}
where \(\mathbf{z}_t^i\) are discrete-action logits and \(V_t^i\) is the state-value estimate. The actor therefore outputs a categorical policy over motion primitives, while the critic provides the value target used by PPO.

\subsection{Cooperative View Policy Learning}
COLMAR uses a shared policy across all agents and environments. This subsection describes policy optimization only. All observation features and reward signals below come from the shared TSDF during training. 3DGS is not used in the policy forward pass or PPO update, and appears only later for photorealistic evaluation of the selected trajectories.
At step $t$, each agent $i$ receives map-centric observation $o_t^i$ and predicts discrete action logits
\begin{equation}
\mathbf{z}_t^i = f_{\theta}(o_t^i), \quad \mathbf{z}_t^i \in \mathbb{R}^{K},
\end{equation}
where \(f_{\theta}(\cdot)\) is the shared policy network with parameters \(\theta\), \(o_t^i\) is agent \(i\)'s observation at step \(t\), \(\mathbf{z}_t^i\) are action logits, and \(K\) is the number of discrete actions. The policy is modeled as a categorical distribution over this action set:
\begin{equation}
\pi_{\theta}(a_t^i \mid o_t^i)
=
\mathrm{Cat}\!\left(\mathbf{z}_t^i\right).
\end{equation}
Here \(\pi_{\theta}(a_t^i \mid o_t^i)\) is the probability of selecting action \(a_t^i\) given \(o_t^i\), and \(\mathrm{Cat}(\cdot)\) denotes a categorical distribution parameterized by logits.
The discrete action space contains motion primitives for translation in $x$, $y$, and $z$, and orientation changes in yaw and pitch. In all experiments, the maximum translation increment per action is \(0.25\) m and the maximum rotation increment per action is \(15^\circ\) (yaw/pitch), with roll disabled.

Actions are sampled independently per agent at deployment with frozen policy parameters. Cooperation emerges from shared map context and from shared policy parameters learned jointly during training.
Compared with peer-to-peer action messaging or a centralized planner, each agent samples its own action independently. Map fusion and overlap-aware rewards provide an implicit coordination signal.

Policy parameters are updated with the standard PPO objective \cite{schulman2017ppo} on trajectories collected from all agents and environments. The importance ratio
\begin{equation}
\rho_t(\theta)=\frac{\pi_\theta(a_t\mid o_t)}{\pi_{\theta_{\text{old}}}(a_t\mid o_t)}
\end{equation}
measures how much the updated policy probability of sampled action \(a_t\) at observation \(o_t\) deviates from the behavior policy \(\pi_{\theta_{\text{old}}}\) used to collect rollouts. Advantages are estimated with generalized advantage estimation:
\begin{equation}
\hat{A}_t=\sum_{l=0}^{T-t-1}(\gamma\lambda)^l\,\delta_{t+l},
\qquad
\delta_t=r_t+\gamma V(o_{t+1})-V(o_t).
\end{equation}
Using \(\rho_t(\theta)\) and \(\hat{A}_t\), we define
\begin{equation}
\begin{aligned}
\mathcal{L}_{1}(\theta) &= \rho_t(\theta)\hat{A}_t, \\
\mathcal{L}_{2}(\theta) &= \mathrm{clip}\!\left(\rho_t(\theta),1-\epsilon,1+\epsilon\right)\hat{A}_t.
\end{aligned}
\end{equation}
The PPO clipped objective is then
\begin{equation}
\mathcal{L}_{\text{clip}}(\theta)=\mathbb{E}_t\!\left[\min\!\left(\mathcal{L}_{1}(\theta),\mathcal{L}_{2}(\theta)\right)\right].
\end{equation}
The final training objective combines this policy term with a value-regression loss and entropy regularization, optimized over shuffled mini-batches for multiple epochs per rollout. This stabilizes updates during training. At deployment, only independent per-agent action selection is required.

\begin{figure}[t]
\centering
\includegraphics[width=0.48\linewidth]{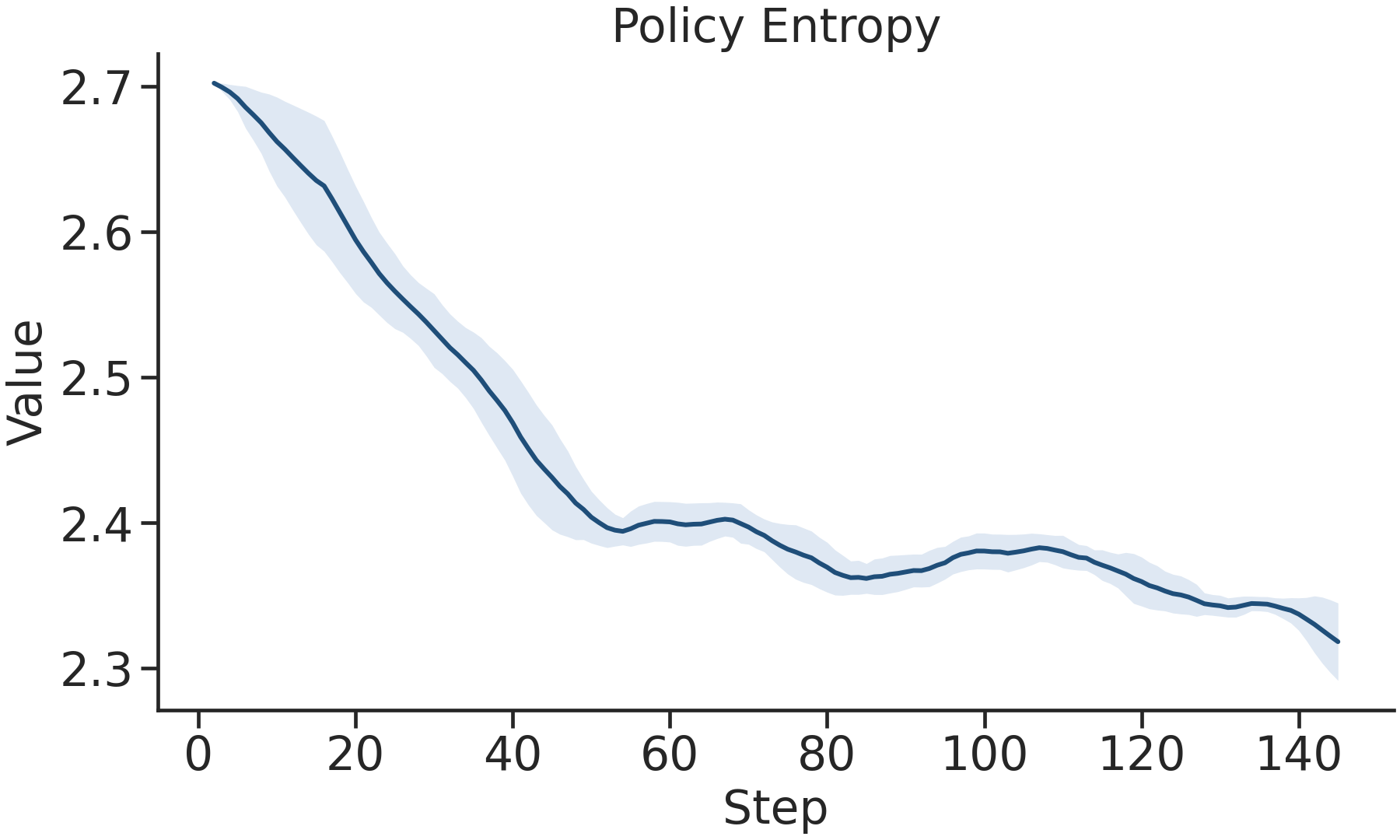}
\hfill
\includegraphics[width=0.48\linewidth]{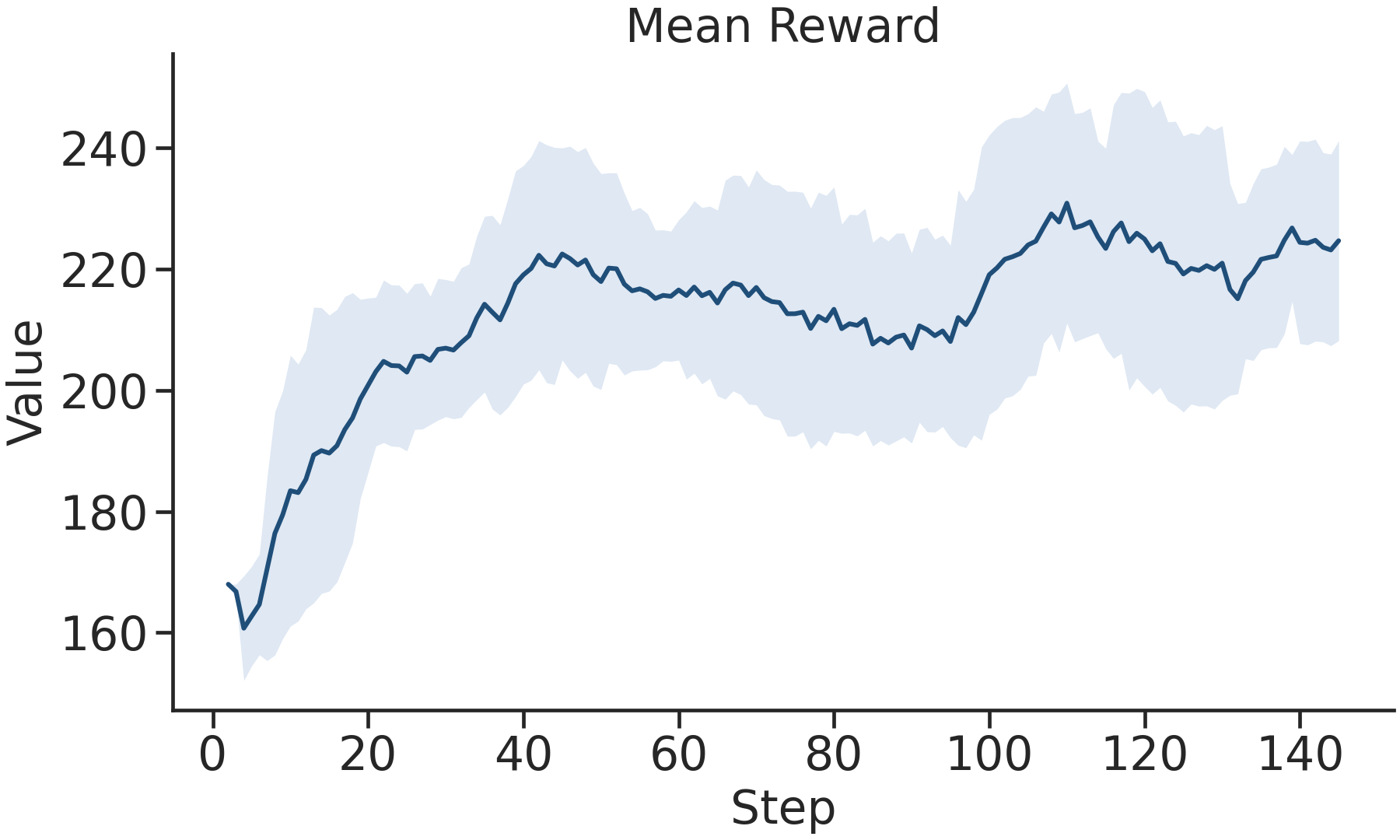}
\includegraphics[width=0.48\linewidth]{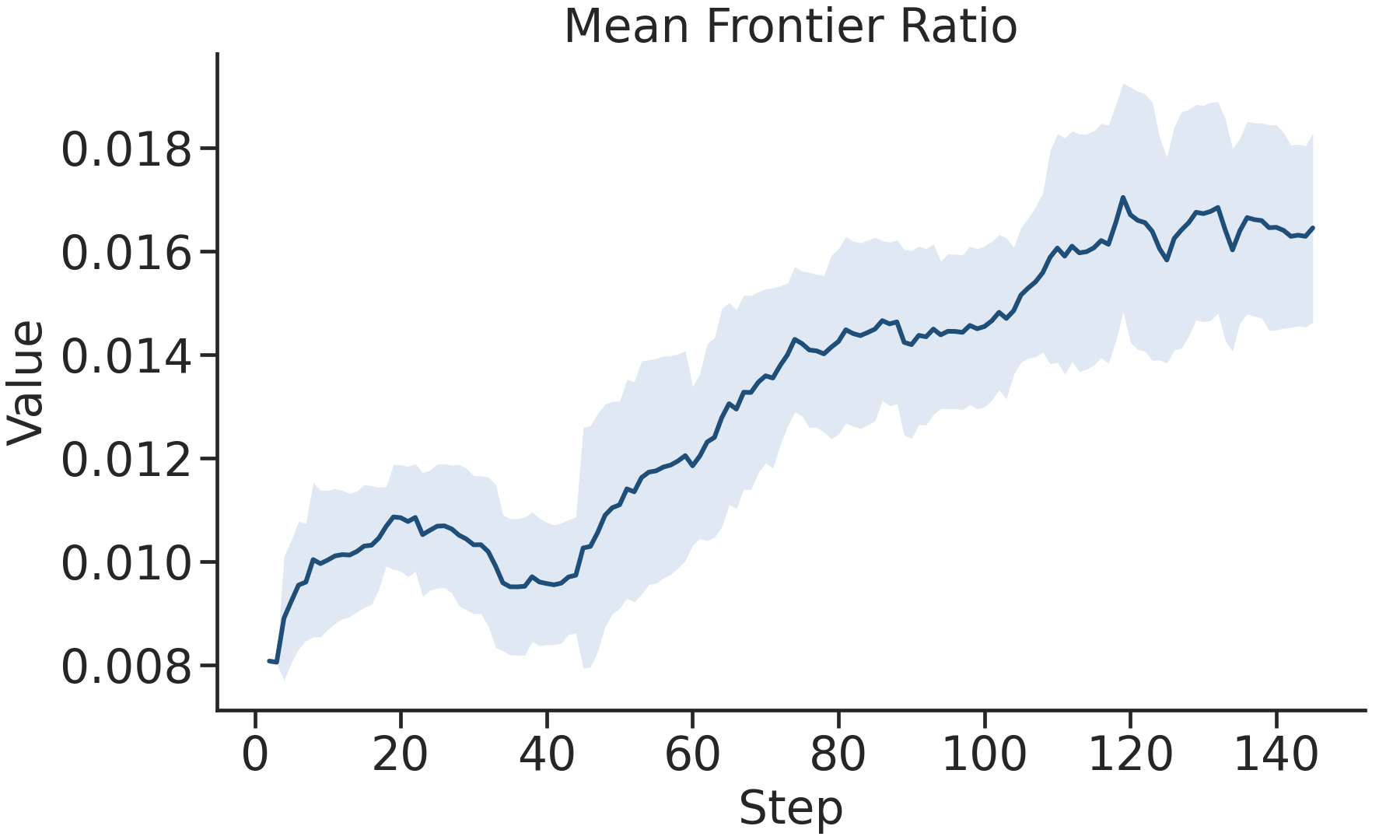}
\hfill
\includegraphics[width=0.48\linewidth]{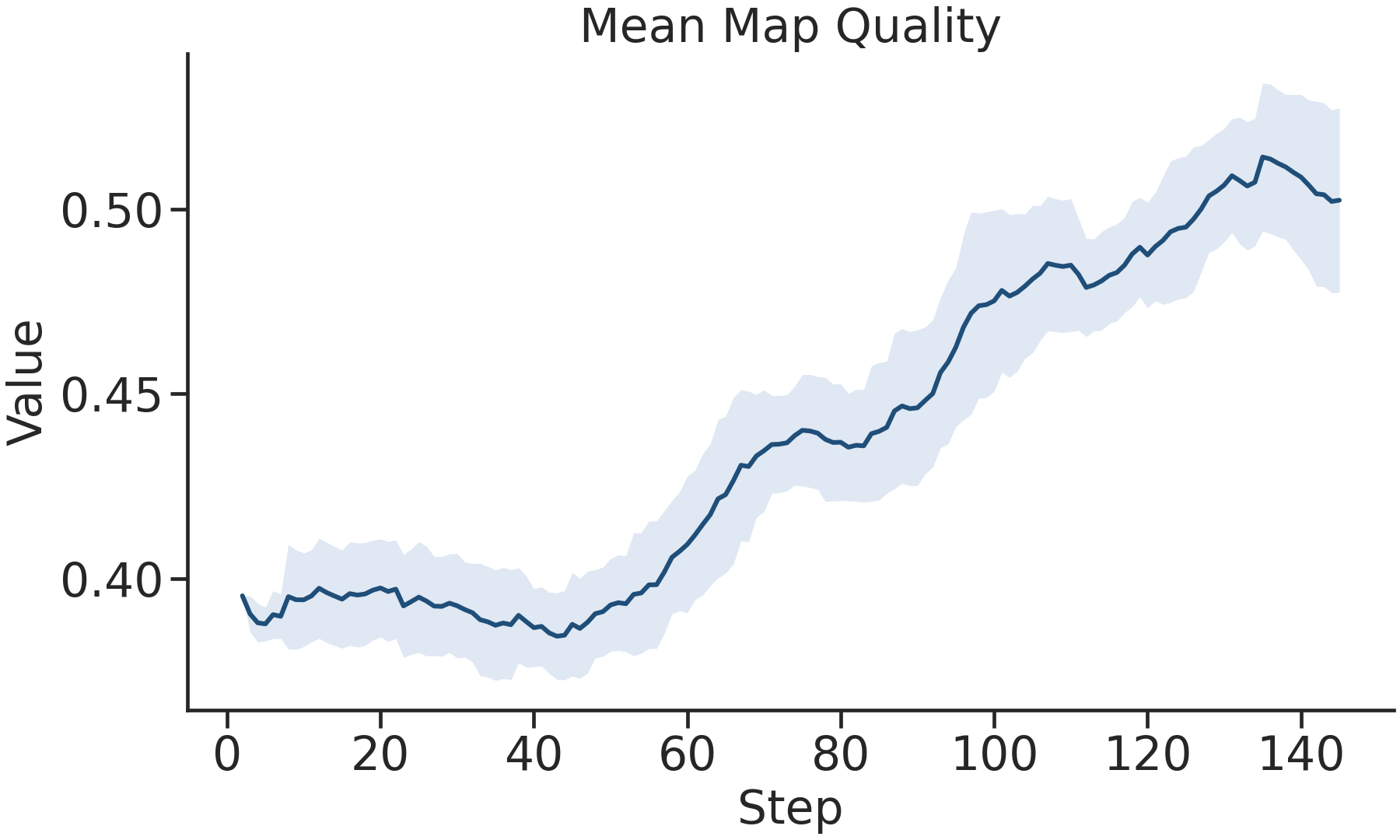}
\caption{Training dynamics of \ours shown as running averages over training rollouts. From left to right, top to bottom: policy entropy, mean episode reward, mean frontier ratio, and mean map quality versus training updates.}
\label{fig:training_dynamics}
\end{figure}

\subsection{Reconstruction-Aware Reward}

\begin{table*}[t]
\centering
\caption{Main quantitative comparison on Replica. Higher is better for Coverage, Accuracy, PSNR, and SSIM. Lower is better for Chamfer, LPIPS, and depth \(L_1\).}
\label{tab:main}
\begin{tabular}{lcccccccc}
\toprule
Method & Coverage (\%) & Accuracy (\%) $\uparrow$ & Chamfer (cm) $\downarrow$ & PSNR $\uparrow$ & SSIM $\uparrow$ & LPIPS $\downarrow$ & Depth $L_1$ (cm) $\downarrow$ \\
\midrule
Random Policy & 55.4 & 58.1 & 14.4 & 25.42 & 0.8375 & 0.1641 & 12.84 \\
Greedy Coverage & 63.9 & 70.7 & 10.7 & 27.88 & 0.9018 & 0.0806 & 7.17 \\
Frontier-Based & 67.6 & 68.0 & 9.30 & 33.24 & 0.8989 & 0.0927 & 5.79 \\
Single-Agent PPO (shared at test) & 74.8 & 77.6 & 6.80 & 30.11 & 0.9132 & 0.0574 & 2.52 \\
\textbf{COLMAR (ours)} & \textbf{82.6} & \textbf{89.4} & \textbf{4.57} & \textbf{37.61} & \textbf{0.9786} & \textbf{0.0312} & \textbf{1.25} \\
\bottomrule
\end{tabular}
\end{table*}

To couple active view planning with downstream reconstruction progress, we shape a dense cooperative reward from incremental TSDF updates. The design encourages each agent to acquire \emph{non-redundant} observations that expand the reconstructed surface while maintaining safe, well-spread team behavior. For readability, we omit the time index \(t\) when the dependence on the current step is clear. Concretely, each agent \(i\) receives
    
\begin{equation}
r^i =
w_{\text{ucov}}\,g_{\text{ucov}}^i
+ b_{\text{team}}
- p_{\text{stag}}^i
- p_{\text{col}}^i
- p_{\text{crowd}}^i .
\end{equation}
Here \(g_{\text{ucov}}^i\) is an overlap-aware \emph{unique} coverage gain and \(b_{\text{team}}\) is a team-level discovery bonus shared across agents. The penalty terms capture different failure modes. \(p_{\text{stag}}^i\) grows when an agent makes little effective progress, as indicated by weak step motion and inefficient recent trajectories (e.g., back-and-forth motion with small net displacement). \(p_{\text{col}}^i\) increases under unsafe contacts, including geometric collisions inferred from map evidence and motion-level collision checks. \(p_{\text{crowd}}^i\) grows when agents cluster too tightly in space, based on nearest-neighbor spacing, to encourage spatial dispersion and reduce redundant sensing.
    
Unique coverage is computed from step-wise coverage growth and a measure of inter-agent overlap:
\begin{align}
g_{\text{lcov}}^i &= \Delta \text{coverage}^i, \\
g_{\text{ucov}}^i &= \frac{g_{\text{lcov}}^i}{\text{overlap}^i}
+ \frac{b_{\text{new}}\,\mathbf{1}_{\text{new-cell}}}{\text{overlap}^i} 
\end{align}
\(\Delta \text{coverage}^i\) denotes the step-wise reconstruction progress attributable to agent \(i\)'s action. In our implementation, we measure this as a view-space gain computed from TSDF renders: after fusing the current depth measurement, a pixel contributes if the rendered depth from the fused TSDF becomes closer than before (i.e., \(d_{\text{after}}(u,v) < d_{\text{before}}(u,v)\)). \(\Delta \text{coverage}^i\) is the fraction of such pixels. \(\text{overlap}^i \) quantifies how many agents concurrently attend to the same region. Dividing by \(\text{overlap}^i\) implements a soft credit assignment that favors complementary viewpoints when multiple agents compete for similar observations. The new-cell bonus \(b_{\text{new}}\mathbf{1}_{\text{new-cell}}\) further emphasizes first-time discovery, biasing exploration toward previously unseen areas.
This ratio form provides a lightweight credit-allocation mechanism: for equal raw gain, rewards decrease as more agents overlap on the same region, so specialization is preferred over redundant co-observation.
    
The team bonus is
\begin{equation}
b_{\text{team}}
=
w_{\text{team}}
\frac{n_{\text{new}}}{n_{\text{alive}}},
\end{equation}
where \(n_{\text{new}}\) is the number of newly discovered cells by the team at the current step, \(n_{\text{alive}}\) is the number of active agents, and \(w_{\text{team}}\) controls the strength of this shared signal. Normalizing by \(n_{\text{alive}}\) keeps reward magnitudes comparable across different team sizes.
    
Together, these terms provide dense feedback for long-horizon coordination by rewarding novel map growth at both individual and team levels while penalizing oscillatory, unsafe, or redundant behaviors. Under static-scene assumptions and consistent TSDF fusion, maximizing incremental view-space gain serves as a practical proxy for reducing unseen surface area and improving geometric reconstruction quality.

Training dynamics further indicate stable optimization. As shown in Figure~\ref{fig:training_dynamics}, policy entropy decreases smoothly while reward and map-quality metrics improve, suggesting a gradual exploration-to-exploitation transition without premature collapse.

\section{EXPERIMENTAL SETUP}
\label{sec:exp_setup}
\subsection{Datasets and Splits}
We train and evaluate the policy primarily on GLEAM \cite{chen2025gleam}. In our setup, GLEAM provides two training splits (512 scenes each) and one evaluation split (128 scenes). The scenes are preprocessed with texture removed following the GLEAM protocol \cite{chen2025gleam}. To test domain transfer, we additionally evaluate zero-shot generalization on Replica \cite{straub2019replica}, a synthetic indoor dataset with 18 scenes, without policy finetuning.

\subsection{Baselines}
We compare \ours against heuristic and learning baselines under matched sensing budgets and action spaces: (1) random action sampling from the same action space, (2) greedy one-step coverage maximization over discrete candidates, (3) frontier-driven exploration using occupancy/frontier maps, and (4) a single-agent PPO policy variant used as a non-cooperative learning baseline. 

\subsection{Implementation Details}
All methods use the same discrete action parameterization (translation in \(x,y,z\), rotation in yaw and pitch) and identical rendering settings. During policy training, a shared TSDF maintained by the mapping backend provides online occupancy and frontier features as well as reconstruction-aware rewards, keeping geometry signals stable inside the RL loop. At deployment, agents observe the same class of map-centric features derived from fused team measurements. The reported photorealistic results use 3DGS for final reconstruction and photometric evaluation.

Training uses PPO with learning rate \(1\times10^{-4}\), clip \(\epsilon=0.1\), \(\gamma=0.99\), \(\lambda=0.95\), entropy coefficient \(0.01\), value coefficient \(0.5\), 2 PPO epochs per update, and 128 rollout steps per update. Per-GPU memory footprint is typically in the \(\sim4\)–\(6\) GB range under our default training configuration.

\subsection{Metrics and Protocol}
We evaluate each method with fixed episode budgets and fixed team size settings. Main multi-agent comparisons use \(N\in\{1,2,4\}\), and scaling experiments sweep \(N=1,2,3,4\). Budget-efficiency analysis sweeps sensing horizon \(T\in\{50,100,150,200,300\}\). For each scene, we run multiple evaluation episodes and report the mean. Primary metrics are mean global coverage gain, accuracy, and Chamfer distance. Final reconstruction quality is evaluated with peak signal-to-noise ratio (PSNR), structural similarity index measure (SSIM), learned perceptual image patch similarity (LPIPS), and depth \(L_1\) error (cm). Accuracy (\%) is defined as the percentage of reconstructed points whose nearest-neighbor distance to ground-truth (GT) points is below threshold \(\tau\). We report the results with \(\tau=5\) cm. Chamfer distance (cm) is the symmetric average of nearest-neighbor distances in both directions, \(d_{\text{CD}}=\tfrac{1}{2}(d_{\text{recon}\rightarrow\text{GT}}+d_{\text{GT}\rightarrow\text{recon}})\), summarizing overall geometric reconstruction error.

\section{RESULTS}
\label{sec:results}
\subsection{Main Quantitative Comparison}
Since GLEAM assets are texture-free, photometric reconstruction metrics (PSNR/SSIM/LPIPS) are not meaningful there. Therefore, Table~\ref{tab:main} reports the main quantitative comparison on Replica, where textured ground-truth appearance is available.

The main comparison on Replica shows that \ours consistently outperforms heuristic baselines and the non-cooperative PPO variant under the same sensing budget. Improvements are evident in reconstruction geometry metrics (higher Accuracy and lower Chamfer), indicating that cooperative view allocation improves both exploration efficiency and final surface quality. Final 3DGS rendering quality also improves, with better PSNR/SSIM and lower LPIPS/$L_{1}$-depth errors. Consistent with the abstract summary, the relative gains reach approximately \(54\%\) in reconstruction accuracy and approximately \(49\%\) in coverage ratio against weaker baselines.
\ours achieves \(89.4\%\) accuracy and \(4.57\) cm Chamfer, improving over single-agent PPO (\(77.6\%\), \(6.8\) cm). It also delivers the strongest photometric quality, with PSNR \(37.61\) and LPIPS \(0.0312\).

\begin{figure*}[t]
\centering
\setlength{\tabcolsep}{2pt}
\renewcommand{\arraystretch}{1.0}
\begin{tabular}{cccc}
& \textbf{Step 10} & \textbf{Step 100} & \textbf{Step 200} \\
\rotatebox[origin=c]{90}{\textit{Occupancy map}} &
\raisebox{-0.5\height}{\includegraphics[width=0.12\linewidth]{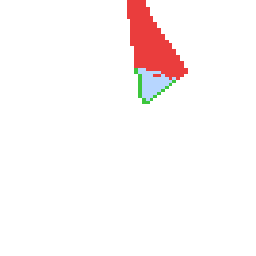}} &
\raisebox{-0.5\height}{\includegraphics[width=0.12\linewidth]{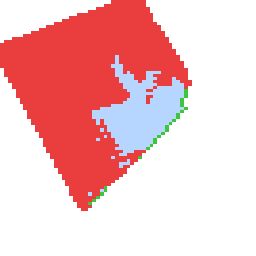}} &
\raisebox{-0.5\height}{\includegraphics[width=0.12\linewidth]{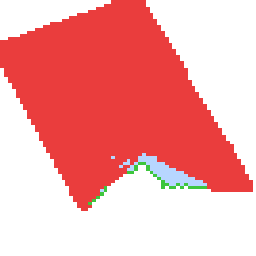}} \\
\rotatebox[origin=c]{90}{\textit{RGB}} &
\raisebox{-0.5\height}{\includegraphics[width=0.33\linewidth]{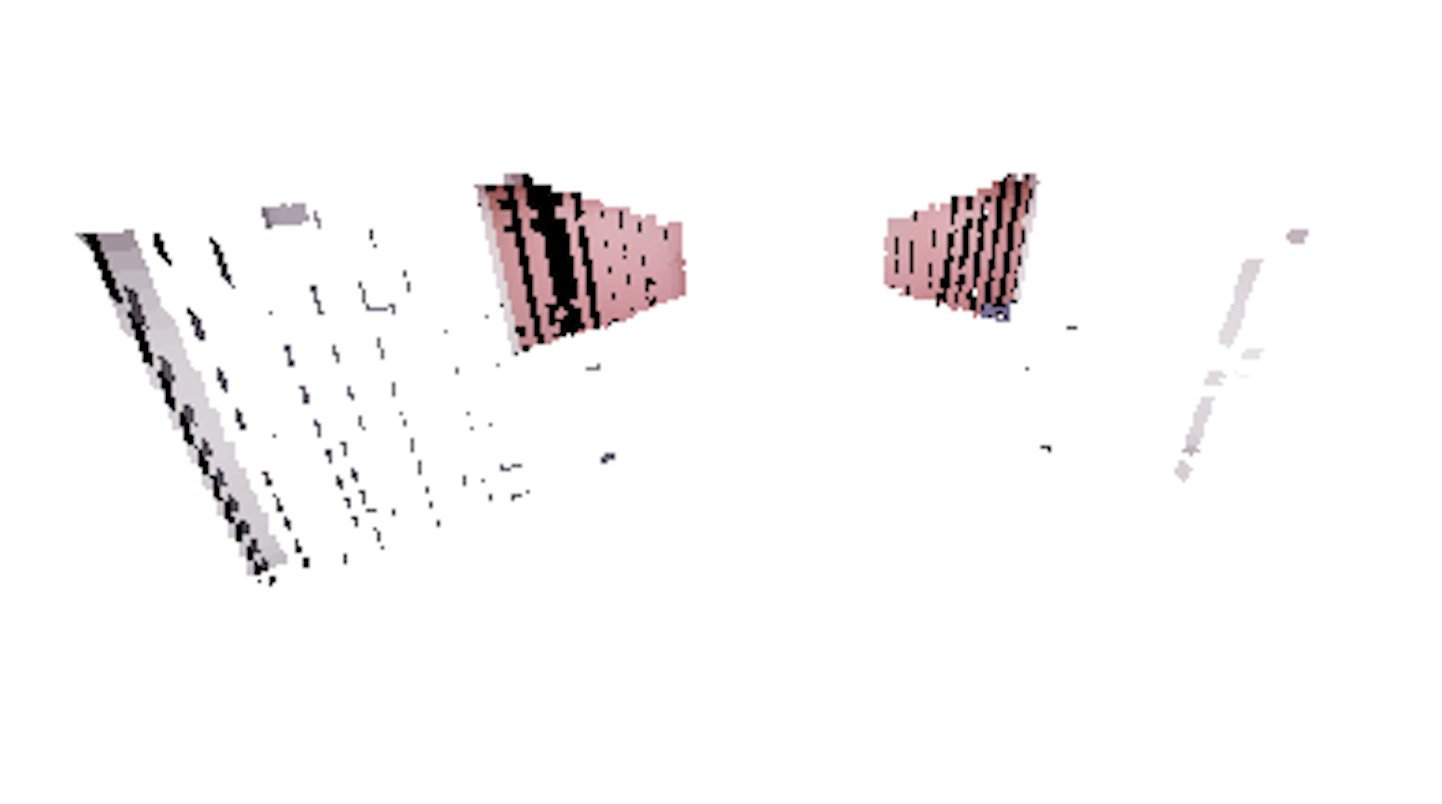}} &
\raisebox{-0.5\height}{\includegraphics[width=0.33\linewidth]{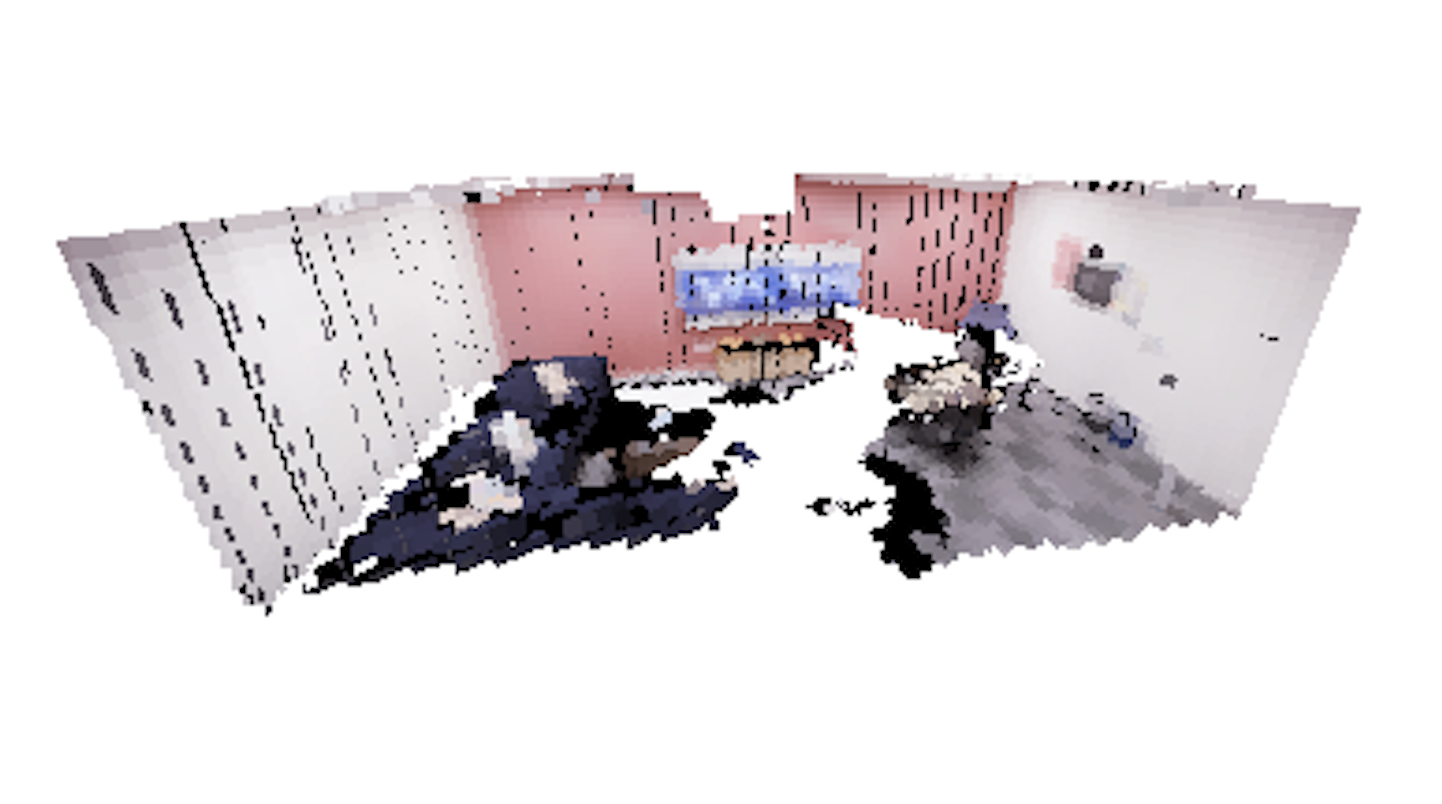}} &
\raisebox{-0.5\height}{\includegraphics[width=0.33\linewidth]{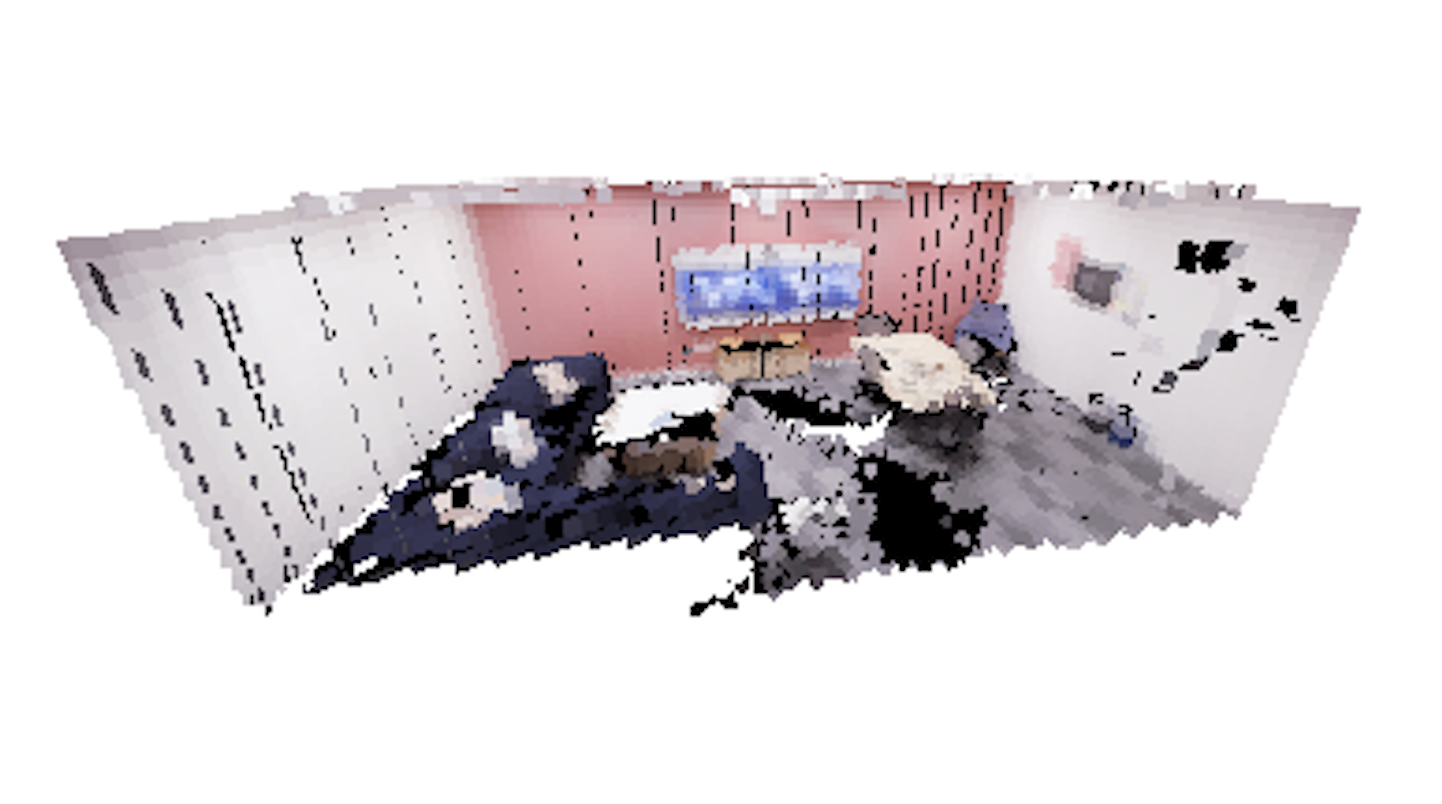}}
\end{tabular}
\caption{Step-wise qualitative reconstruction progress under COLMAR at Steps 10, 100, and 200. Occupancy maps show explored space (red: occupied, light blue: free, green: frontier, white: unknown), and the corresponding RGB renderings are produced from the fused TSDF volume. As exploration proceeds, known regions expand and reconstructions become more complete and consistent. Black patches are rendering artifacts and do not reflect reconstruction errors.}
\label{fig:qual_progress}
\end{figure*}

\subsection{Scaling With Number of Agents}
\begin{table}[t]
\centering
\caption{Scaling with team size on GLEAM.}
\label{tab:scaling}
\begin{tabular}{cccc}
\toprule
\# Agents & Coverage (\%) & Accuracy (\%) $\uparrow$ & Chamfer (cm) $\downarrow$ \\
\midrule
1 & 72.4 & 82.2 & 6.48 \\
2 & 77.1 & 85.2 & 5.61 \\
3 & 79.4 & 86.8 & 5.22 \\
4 & 80.9 & 88.2 & 4.96 \\
\bottomrule
\end{tabular}
\end{table}

As the number of agents increases, performance improves consistently: coverage rises from \(72.4\%\) (\(N=1\)) to \(80.9\%\) (\(N=4\)), and Chamfer drops from \(6.48\) cm to \(4.96\) cm. Gains are strongest at small team sizes and then begin to saturate.

\subsection{Quality-Efficiency Trade-off}
\begin{table}[t]
\centering
\caption{Budget-efficiency trend for \ours on GLEAM.}
\label{tab:budget}
\begin{tabular}{cccc}
\toprule
Budget \(T\) & Coverage (\%) & Accuracy (\%) $\uparrow$ & Chamfer (cm) $\downarrow$ \\
\midrule
50  & 65.1 & 78.1 & 7.95 \\
100 & 72.2 & 82.7 & 6.48 \\
150 & 76.6 & 85.4 & 5.72 \\
200 & 79.1 & 86.9 & 5.25 \\
300 & 80.9 & 88.2 & 4.96 \\
\bottomrule
\end{tabular}
\end{table}

Under matched budget sweeps, \ours shows clear efficiency gains: from \(T=50\) to \(T=300\), coverage improves from \(65.1\%\) to \(80.9\%\) and Chamfer drops from \(7.95\) cm to \(4.96\) cm. Improvements remain monotonic but slow down at larger budgets.

\subsection{Cross-Dataset Generalization}

Training on GLEAM and evaluating zero-shot on Replica demonstrates good transfer. Replica is slightly better than GLEAM (e.g., \(4.57\) vs \(4.96\) cm Chamfer), consistent with simpler scene geometry.

\subsection{Qualitative Results}

Qualitative results confirm the coordination behavior learned by \ours. In Figure~\ref{fig:qual_progress}, the occupancy maps (top row) show steady growth of known free/occupied space and shrinking frontier pockets from Step 10 to Step 200, while the rendered reconstructions (bottom row) become more complete with fewer visible holes and artifacts. Compared with non-cooperative trajectories, agents in \ours spread to complementary frontiers instead of repeatedly revisiting the same local region, which improves both map growth and final rendering consistency. Failure cases are mostly observed in cluttered corners and thin-structure regions, where occlusion and long-horizon credit assignment still limit view allocation quality.

\begin{figure*}[t]
\centering
\setlength{\tabcolsep}{2pt}
\renewcommand{\arraystretch}{1.0}
\begin{tabular}{ccccccc}
& \multicolumn{2}{c}{\textbf{Agent 1}} & \multicolumn{2}{c}{\textbf{Agent 2}} & \multicolumn{2}{c}{\textbf{Agent 3}} \\
\rotatebox[origin=c]{90}{\textit{GT RGB}} &
\raisebox{-0.5\height}{\includegraphics[width=0.145\linewidth]{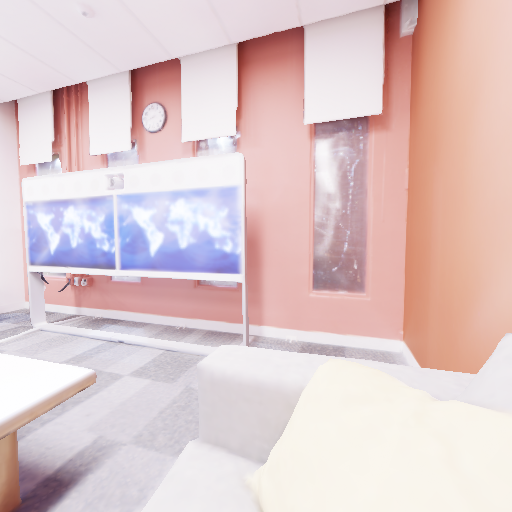}} &
\raisebox{-0.5\height}{\includegraphics[width=0.145\linewidth]{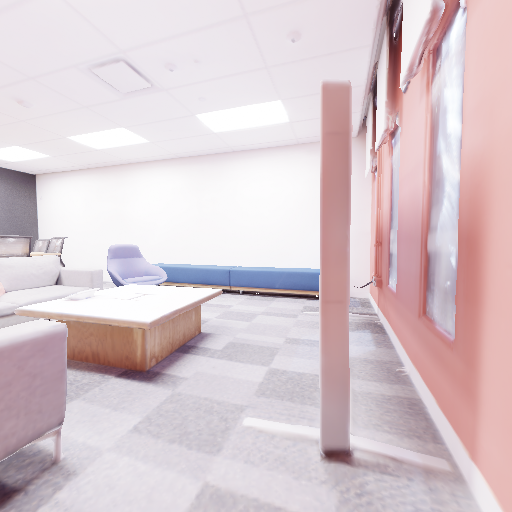}} &
\raisebox{-0.5\height}{\includegraphics[width=0.145\linewidth]{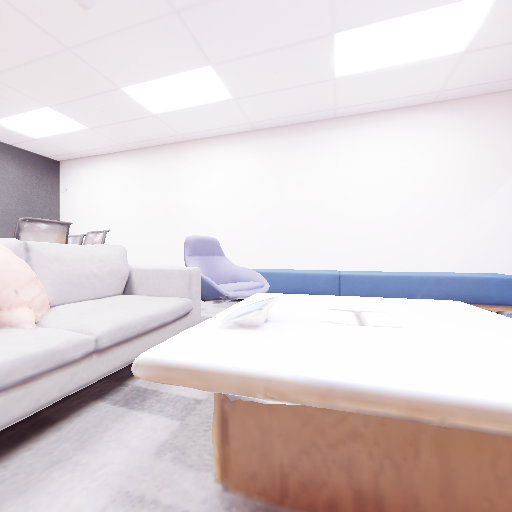}} &
\raisebox{-0.5\height}{\includegraphics[width=0.145\linewidth]{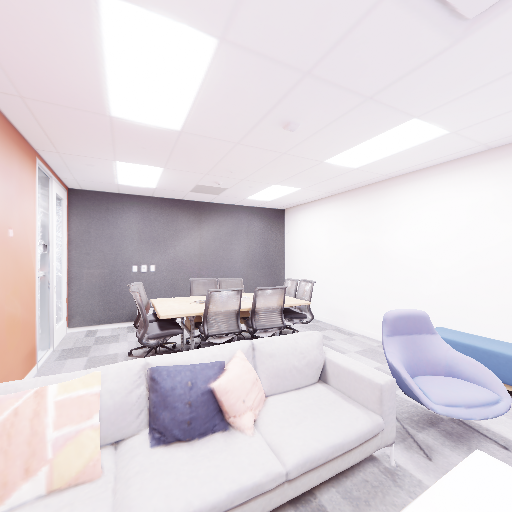}} &
\raisebox{-0.5\height}{\includegraphics[width=0.145\linewidth]{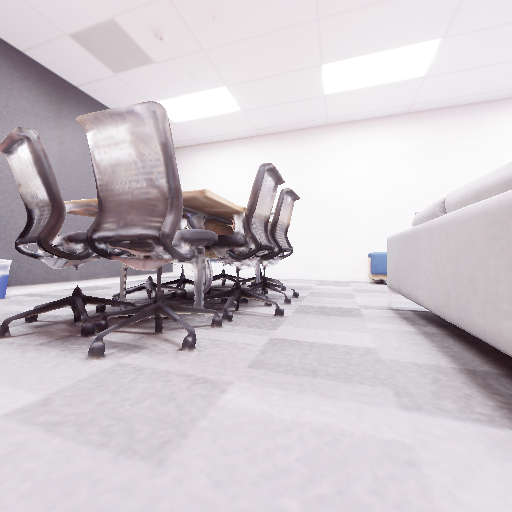}} &
\raisebox{-0.5\height}{\includegraphics[width=0.145\linewidth]{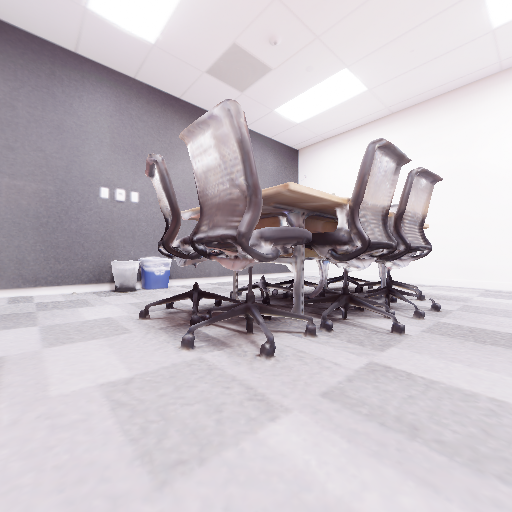}} \\
\rotatebox[origin=c]{90}{\textit{GT Depth}} &
\raisebox{-0.5\height}{\includegraphics[width=0.145\linewidth]{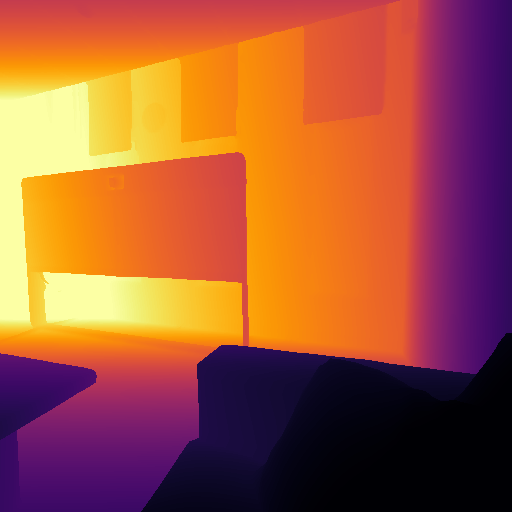}} &
\raisebox{-0.5\height}{\includegraphics[width=0.145\linewidth]{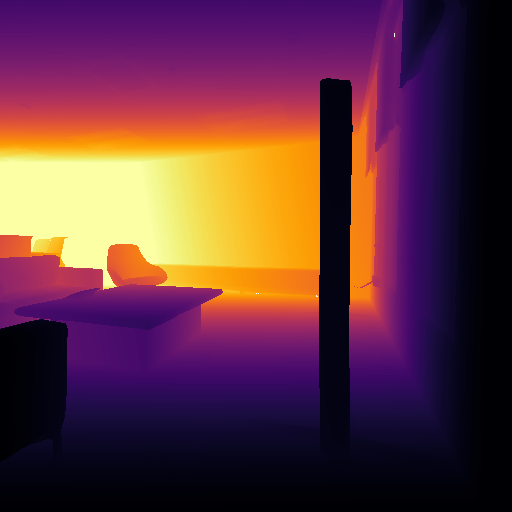}} &
\raisebox{-0.5\height}{\includegraphics[width=0.145\linewidth]{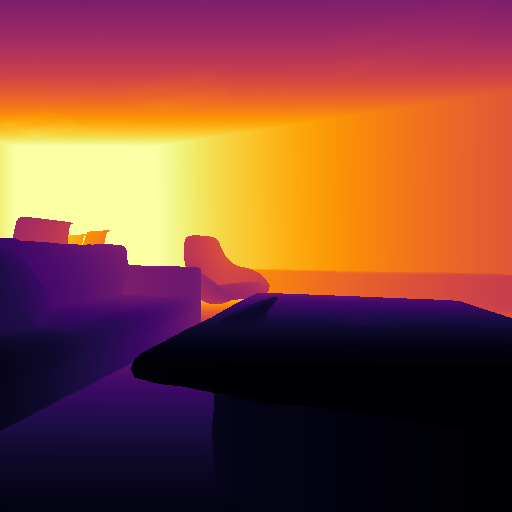}} &
\raisebox{-0.5\height}{\includegraphics[width=0.145\linewidth]{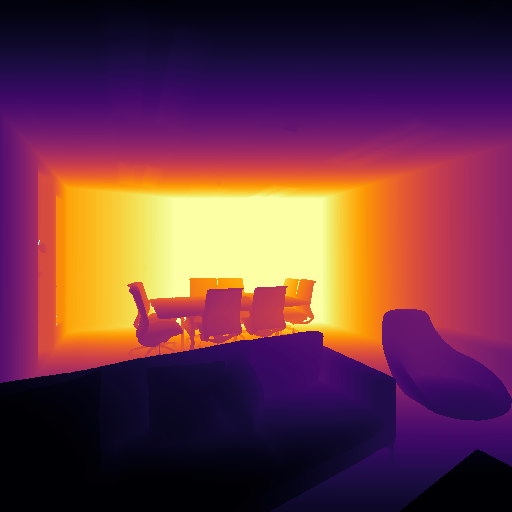}} &
\raisebox{-0.5\height}{\includegraphics[width=0.145\linewidth]{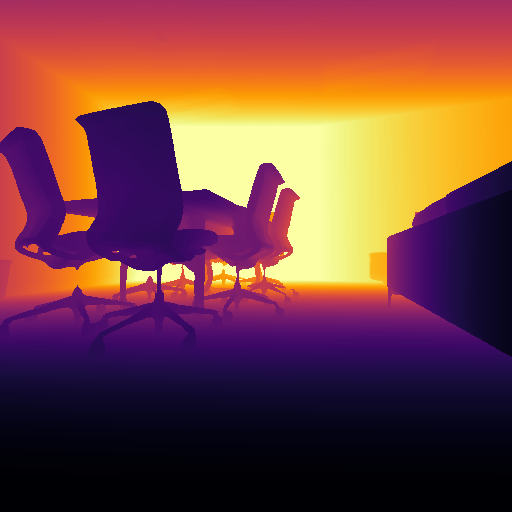}} &
\raisebox{-0.5\height}{\includegraphics[width=0.145\linewidth]{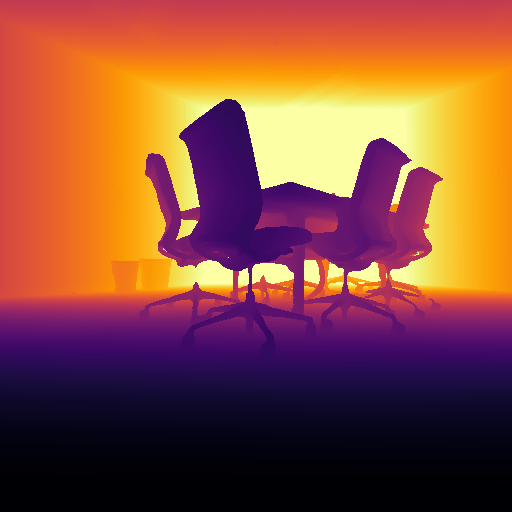}} \\
\rotatebox[origin=c]{90}{\textit{Recon RGB}} &
\raisebox{-0.5\height}{\includegraphics[width=0.145\linewidth]{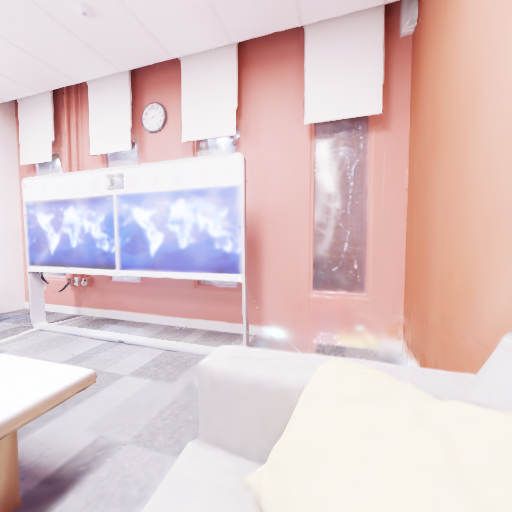}} &
\raisebox{-0.5\height}{\includegraphics[width=0.145\linewidth]{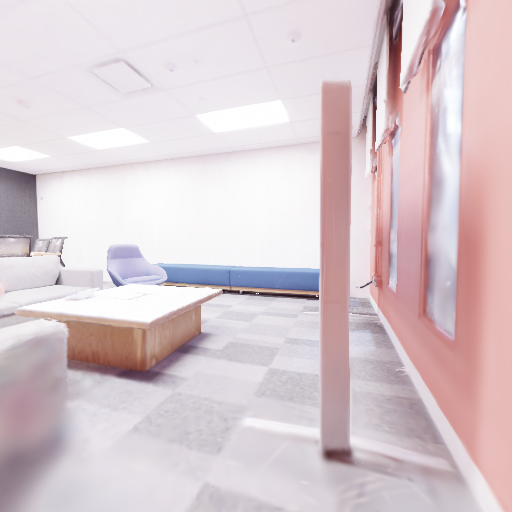}} &
\raisebox{-0.5\height}{\includegraphics[width=0.145\linewidth]{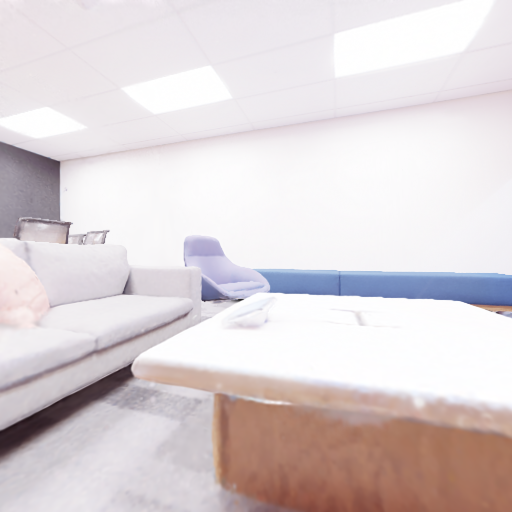}} &
\raisebox{-0.5\height}{\includegraphics[width=0.145\linewidth]{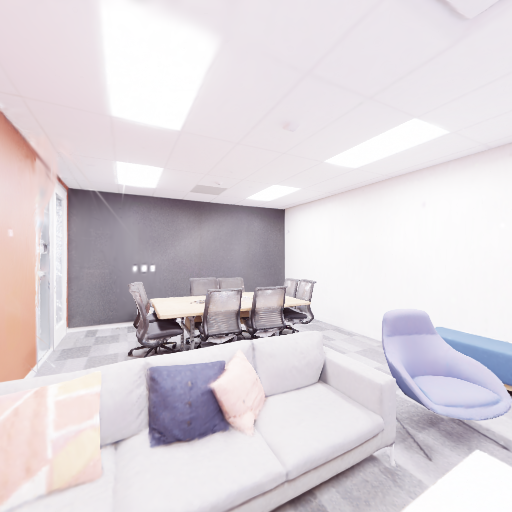}} &
\raisebox{-0.5\height}{\includegraphics[width=0.145\linewidth]{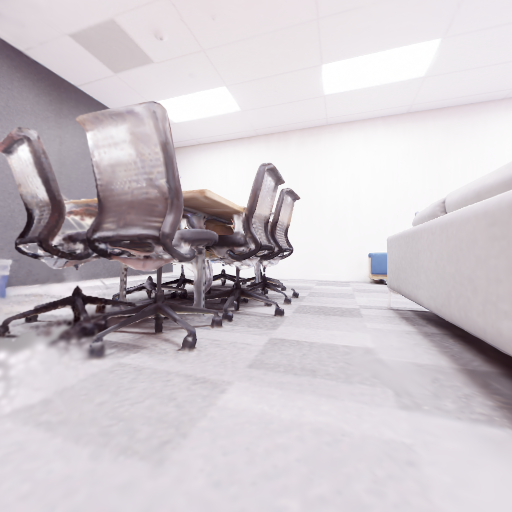}} &
\raisebox{-0.5\height}{\includegraphics[width=0.145\linewidth]{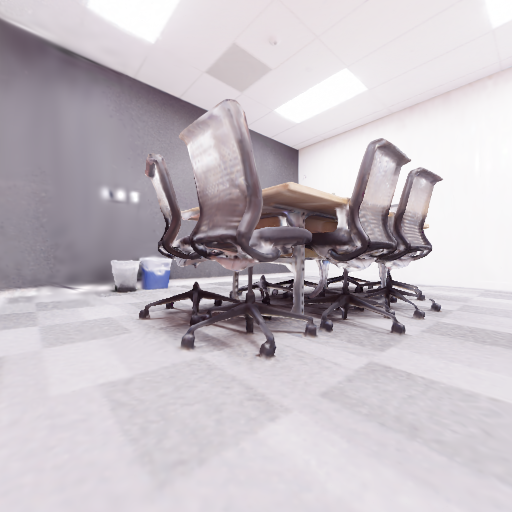}} \\
\rotatebox[origin=c]{90}{\textit{Recon Depth}} &
\raisebox{-0.5\height}{\includegraphics[width=0.145\linewidth]{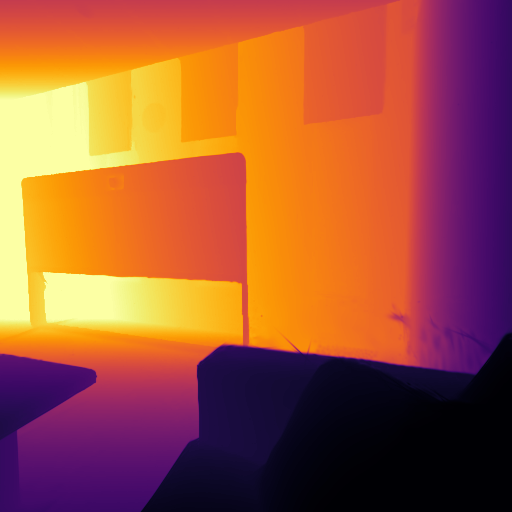}} &
\raisebox{-0.5\height}{\includegraphics[width=0.145\linewidth]{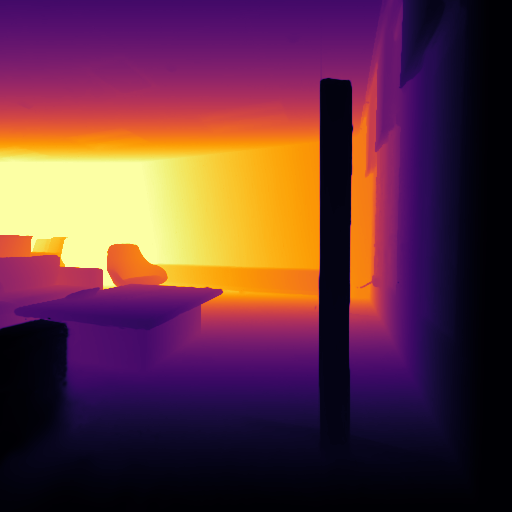}} &
\raisebox{-0.5\height}{\includegraphics[width=0.145\linewidth]{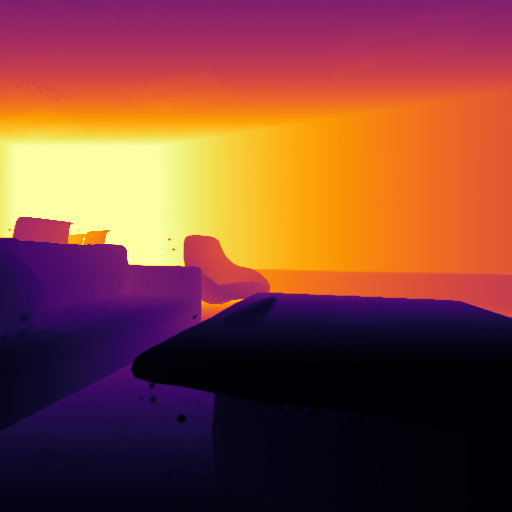}} &
\raisebox{-0.5\height}{\includegraphics[width=0.145\linewidth]{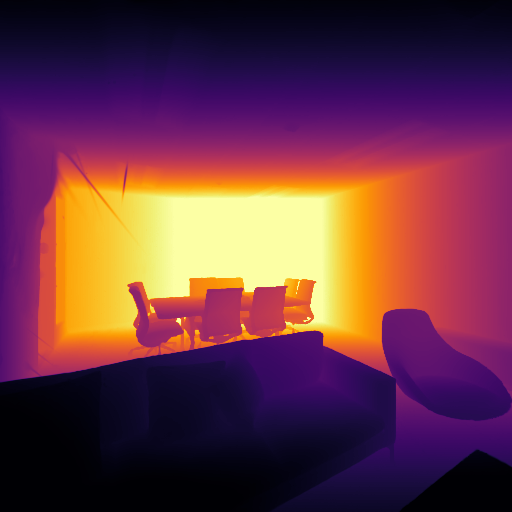}} &
\raisebox{-0.5\height}{\includegraphics[width=0.145\linewidth]{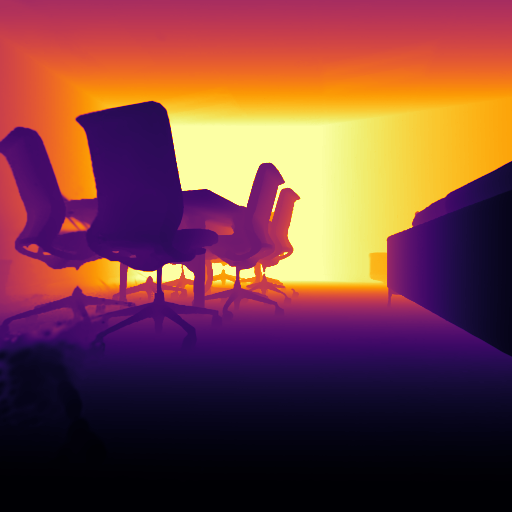}} &
\raisebox{-0.5\height}{\includegraphics[width=0.145\linewidth]{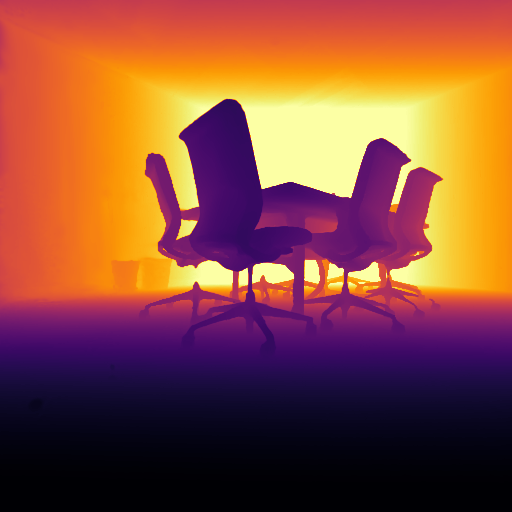}}
\end{tabular}
\caption{Agent-wise qualitative reconstruction under \ours (two snapshots per agent). Columns are grouped by agent to visualize role specialization. The grouped views show that cooperative planning yields spatially complementary trajectories: agents spread out over distinct scene regions, which reduces inter-agent redundancy and improves reconstruction completeness.}
\label{fig:qual_3dgs}
\end{figure*}

Figure~\ref{fig:qual_3dgs} further indicates the coordination pattern induced by \ours. Rather than collapsing to overlapping trajectories, agents distribute themselves across different parts of the scene and repeatedly observe their assigned regions from complementary angles. This spatial specialization improves local surface consolidation while preserving team-level coverage diversity, leading to stronger agreement between reconstructed RGB/depth and GT across agent groups.

\section{ABLATION STUDIES}
\label{sec:ablation}
We conduct ablation studies to understand the effect of each component and justify the key design choices. All ablations follow the same protocol as the main experiments for fair comparison, including identical dataset split, sensing budget, episode horizon, action space, and optimization schedule. Each variant modifies only one factor under study (reward component, training regime, or input scope) and is evaluated under the same multi-agent settings (\(N\in\{1,2,4\}\)) and seed protocol. We report accuracy, final coverage, and Chamfer distance, with collision statistics additionally reported for reward-component analysis.

\begin{table}[t]
\centering
\caption{Cross-dataset transfer from GLEAM to Replica.}
\label{tab:transfer}
\begin{tabular}{lccc}
\toprule
Setting & Coverage $\uparrow$ & Accuracy $\uparrow$ & Chamfer $\downarrow$ \\
        & (\%) & (\%) & (cm) \\
\midrule
In-domain (GLEAM eval) & 80.9 & 88.2 & 4.96 \\
Zero-shot (Replica) & 82.6 & 89.4 & 4.57 \\
\bottomrule
\end{tabular}
\end{table}

\subsection{Reward Components}
Starting from the full reward, we remove one component at a time and retrain with the same policy architecture and PPO schedule.

\begin{table}[t]
\centering
\caption{Reward component ablation.}
\label{tab:ablation_reward}
\begin{tabular}{lcccc}
\toprule
Variant & Accuracy & Coverage & Chamfer & Collision \\
        & (\%) & (\%) & (cm) & Rate \\
\midrule
\textbf{Full reward} & \textbf{88.2} & \textbf{80.9} & \textbf{4.96} & \textbf{0.07} \\
- unique coverage & 86.8 & 79.0 & 5.33 & 0.08 \\
- team bonus & 87.4 & 79.6 & 5.19 & 0.08 \\
- all penalties & 82.1 & 74.9 & 6.31 & 0.14 \\
\bottomrule
\end{tabular}
\end{table}

This study isolates the contribution of cooperative utility and regularization terms in the reward. The full reward performs best (\(88.2\%\) accuracy and \(4.96\) cm Chamfer distance), while removing all penalties causes the largest degradation (\(82.1\%\), \(6.31\) cm) and doubles collision rate (from \(0.07\) to \(0.14\)).

\subsection{Cooperative vs Non-Cooperative Training}
The cooperative variant is trained jointly with multi-agent rollouts, whereas the non-cooperative baseline is trained in a single-agent setting and then reused independently by all agents at test time.
\begin{table}[t]
\centering
\caption{Cooperative vs non-cooperative training.}
\label{tab:ablation_coop}
\begin{tabular}{lccc}
\toprule
Training Scenario & Accuracy (\%) & Cov. (\%) & Chamfer (cm) \\
\midrule
Single-agent & 84.5 & 73.8 & 5.89 \\
\textbf{Multi-agent} & \textbf{88.2} & \textbf{80.9} & \textbf{4.96} \\
\bottomrule
\end{tabular}
\end{table}
This comparison quantifies the effect of joint multi-agent optimization. Multi-agent training improves both geometry and coverage, reducing Chamfer from \(5.89\) cm to \(4.96\) cm and increasing coverage from \(73.8\%\) to \(80.9\%\).

\subsection{Global vs Local Input Ablation}
We keep the same backbone and training schedule, and mask input channels to retain only local cues, only global cues, or both.
\begin{table}[t]
\centering
\caption{Global vs local input ablation.}
\label{tab:ablation_modality}
\begin{tabular}{lccc}
\toprule
Input scope to policy & Accuracy & Coverage & Chamfer \\
& (\%) & (\%) & (cm) \\
\midrule
Depth + Ego Occupancy (local) & 86.7 & 79.2 & 5.35 \\
Pose-grid/map cues (global) & 85.9 & 78.3 & 5.58 \\
\textbf{Full (local + global)} & \textbf{88.2} & \textbf{80.9} & \textbf{4.96} \\
\bottomrule
\end{tabular}
\end{table}
The full input set performs best (\(88.2\%\), \(4.96\) cm). Local-only input degrades performance, and global-only is worst (\(5.58\) cm), confirming that local geometry and global coordination cues are complementary.

\section{DISCUSSION AND LIMITATIONS}
\label{sec:discussion}

COLMAR is most effective in static indoor scenes where cooperative view allocation can exploit complementary visibility under fixed sensing budgets, improving geometric quality and exploration efficiency over heuristic and non-cooperative baselines through overlap-aware rewards and shared policy learning.

Nevertheless, the current study is limited to static environments; dynamic changes, sensor noise, calibration errors, and domain shift may degrade performance. Failure cases arise in narrow or heavily occluded layouts, and gains saturate as team size increases. Real-world deployment additionally requires robust multi-agent localization, reliable communication, and safety-aware low-level control.

\section{CONCLUSION}
\label{sec:conclusion}
We presented COLMAR, a cooperative view policy learning framework for multi-agent active 3D reconstruction. The method combines map-centric policy inputs with reconstruction-aware reward design to promote complementary exploration, reduce redundancy, and improve reconstruction quality under limited sensing budgets. Across benchmark evaluations, COLMAR consistently outperforms heuristic and learning baselines in both geometric and photometric metrics, and shows favorable scaling with team size and sensing budget. Future work can investigate real-robot deployment with improved robustness to sensor noise and calibration drift, communication-aware coordination under bandwidth constraints, and extensions to dynamic scenes with temporally consistent mapping and reward design.

                                  % ====> This causes the references to break in parts

%%%%%%%%%%%%%%%%%%%%%%%%%%%%%%%%%%%%%%%%%%%%%%%%%%%%%%%%%%%%%%%%%%%%%%%%%%%%%%%%

%%%%%%%%%%%%%%%%%%%%%%%%%%%%%%%%%%%%%%%%%%%%%%%%%%%%%%%%%%%%%%%%%%%%%%%%%%%%%%%%

%%%%%%%%%%%%%%%%%%%%%%%%%%%%%%%%%%%%%%%%%%%%%%%%%%%%%%%%%%%%%%%%%%%%%%%%%%%%%%%%
%\section*{APPENDIX}


%\section*{ACKNOWLEDGMENT}

% The preferred spelling of the word ÒacknowledgmentÓ in America is without an ÒeÓ after the ÒgÓ. Avoid the stilted expression, ÒOne of us (R. B. G.) thanks . . .Ó  Instead, try ÒR. B. G. thanksÓ. Put sponsor acknowledgments in the unnumbered footnote on the first page.

%%%%%%%%%%%%%%%%%%%%%%%%%%%%%%%%%%%%%%%%%%%%%%%%%%%%%%%%%%%%%%%%%%%%%%%%%%%%%%%%

{\small
\bibliographystyle{IEEEtran}
\bibliography{refs}
}

\end{document}